\documentclass[10pt,twocolumn,letterpaper]{article}

\usepackage{cvpr}
\usepackage{times}
\usepackage{epsfig}
\usepackage{graphicx}
\usepackage{amsmath}
\usepackage{amssymb}
\usepackage{cuted}
\usepackage{capt-of}
\usepackage{booktabs}
\usepackage[utf8]{inputenc}
\usepackage{tabularx}
\usepackage{listings}
\usepackage{xcolor}
\usepackage{afterpage}
\usepackage{float}
\usepackage{caption}

\definecolor{codegreen}{rgb}{0,0.6,0}
\definecolor{codegray}{rgb}{0.5,0.5,0.5}
\definecolor{codepurple}{rgb}{0.58,0,0.82}
\definecolor{backcolour}{rgb}{0.95,0.95,0.92}
 
\lstdefinestyle{mystyle}{
    backgroundcolor=\color{backcolour},   
    commentstyle=\color{codegreen},
    keywordstyle=\color{magenta},
    numberstyle=\tiny\color{codegray},
    stringstyle=\color{codepurple},
    basicstyle=\ttfamily\footnotesize,
    breakatwhitespace=false,         
    breaklines=true,                 
    captionpos=b,                    
    keepspaces=true,                 
    numbers=left,                    
    numbersep=5pt,                  
    showspaces=false,                
    showstringspaces=false,
    showtabs=false,                  
    tabsize=2
}
 
\usepackage[pagebackref=true,breaklinks=true,letterpaper=true,colorlinks,bookmarks=false]{hyperref}

\usepackage[pagebackref=true,breaklinks=true,letterpaper=true,colorlinks,bookmarks=false]{hyperref}

\cvprfinalcopy 

\def\cvprPaperID{5070\vspace{-1cm}} 

\ifcvprfinal\fi
\begin{document}

\title{Wish You Were Here: Context-Aware Human Generation}

\author{Oran Gafni\\
Facebook AI Research \\
{\tt\small oran@fb.com}
\and Lior Wolf\\
Facebook AI Research and Tel-Aviv University\\
{\tt\small wolf@fb.com}}

\maketitle
\begin{strip}\centering
\begin{tabular}{@{}c@{~}c@{~}c@{~}c@{~}c} 
  \includegraphics[width=0.2\textwidth]{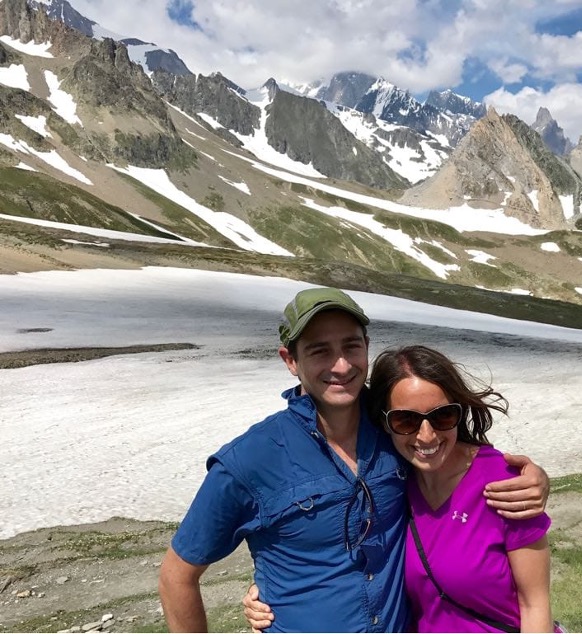} & \includegraphics[width=0.2\textwidth]{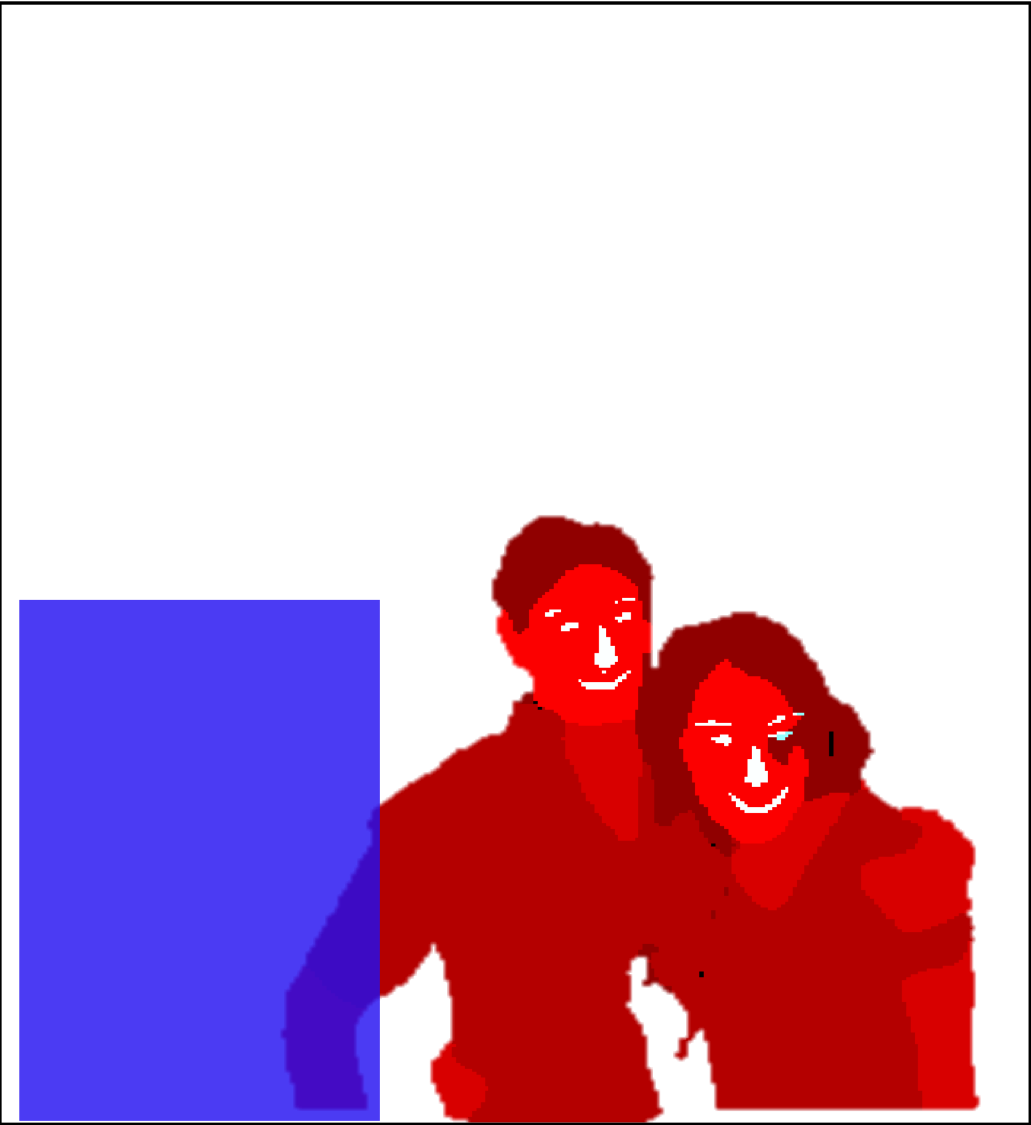} &
  \includegraphics[width=0.2\textwidth]{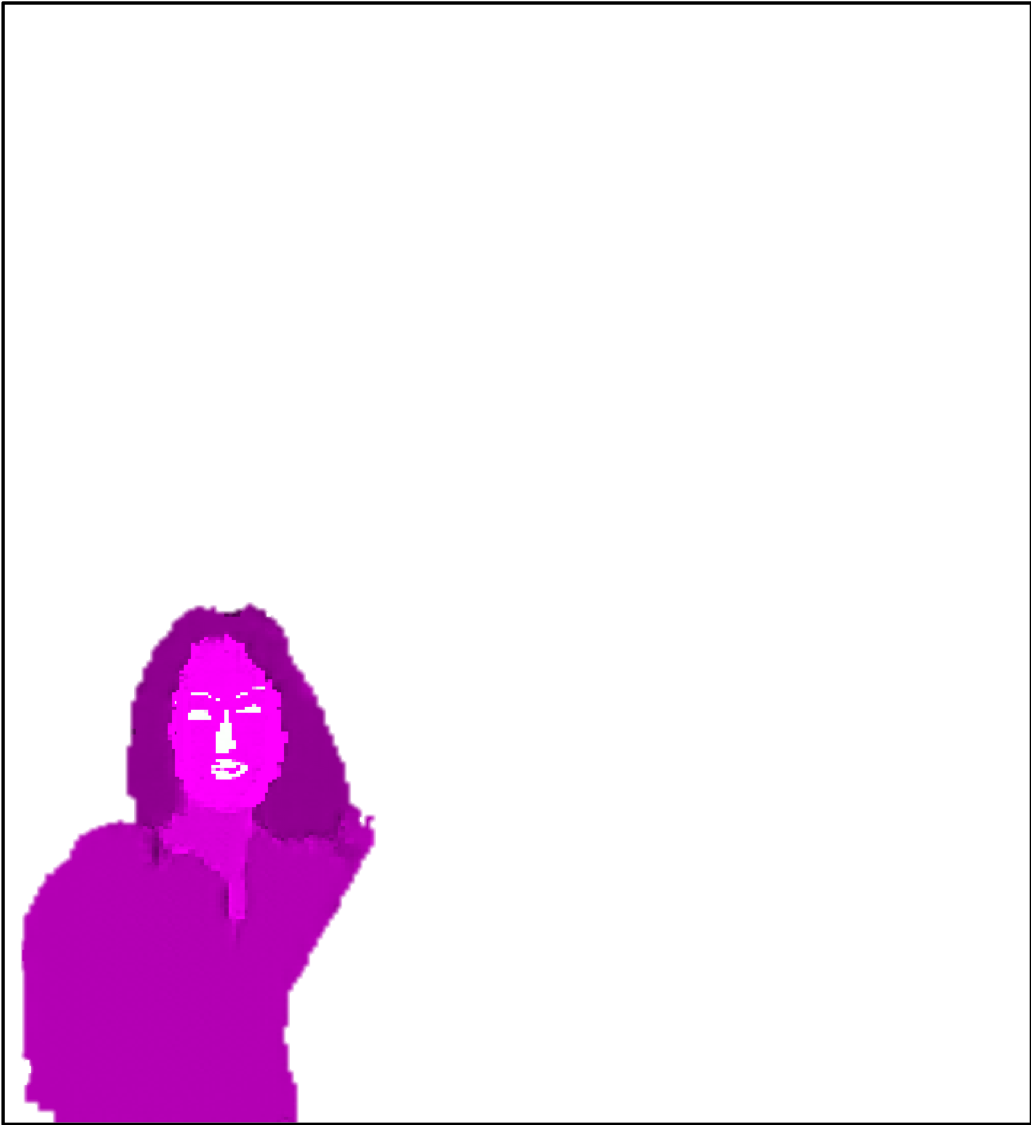} &
  \includegraphics[width=0.065643520982\textwidth]{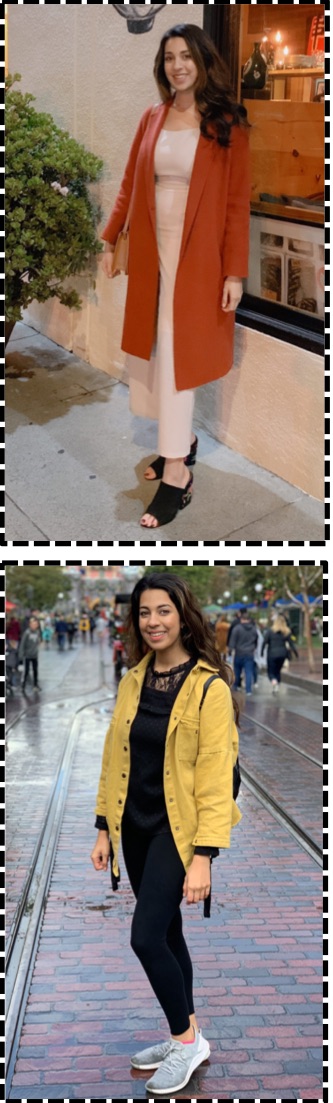} &
  \includegraphics[width=0.135\textwidth]{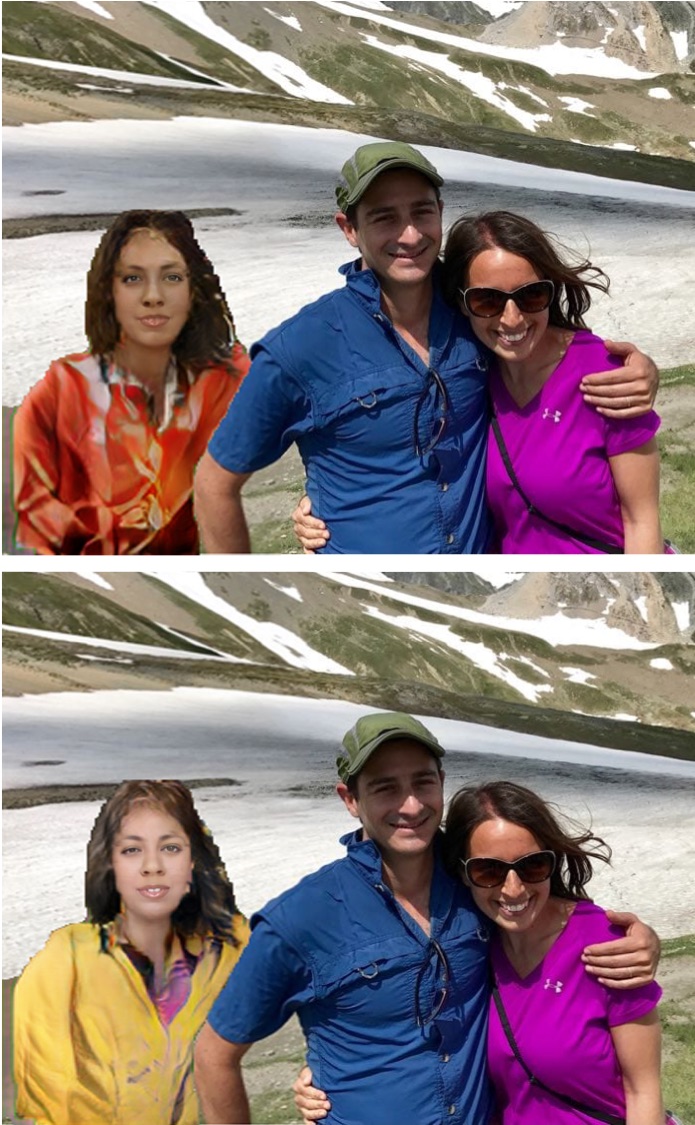} \\
  (a) & (b) & (c) & (d) & (e)
  \end{tabular}

\captionof{figure}{The ``wish you were here'' application. Given an image with one or more persons (a), and an optional bounding box indicating where to add a new person (b), the method generates the pose of the new person (c). {Then, given the appearance information of the target person (d), the method renders a new image (e) with that person.}}
\label{fig:teaser}

\end{strip}

\begin{abstract}
We present a novel method for inserting objects, specifically humans, into existing images, such that they blend in a photorealistic manner, while respecting the semantic context of the scene. Our method involves three subnetworks: the first generates the semantic map of the new person, given the pose of the other persons in the scene and an optional bounding box specification. The second network renders the pixels of the novel person and its blending mask, based on specifications in the form of multiple appearance components. A third network refines the generated face in order to match those of the target person. Our experiments present convincing high-resolution outputs in this novel and challenging application domain. In addition, the three networks are evaluated individually, demonstrating for example, state of the art results in pose transfer benchmarks.
\end{abstract}

\section{Introduction}

The field of image generation has been rapidly progressing in recent years due to the advent of GANs, as well as the introduction of sophisticated architectures and training methods. However, the generation is either done while giving the algorithm an ``artistic freedom'' to generate attractive images, or while specifying concrete constraints such as an approximate drawing, or desired keypoints.

In other contributions, there is a set of semantic specifications, such as in image generation based on scene graphs, or on free text{\color{black}, yet these are not demonstrated to generate high-fidelity human images}. What seems to be missing, is the middle ground of the two: a method that allows some freedom, while requiring adherence to high-level constraints that arise from the image context.

In our work, the generated image has to comply with the soft requirement to have a coherent composition. Specifically, we generate a human figure that fits into the existing scene. Unlike previous work in the domain of human placement, we do not require a driving pose or a semantic map to render a novel person, but rather, we generate a semantic map independently, such that it is suitable to the image context. In addition, we provide rich control over the rendering aspects, enabling additional applications, such as individual component replacement and sketching a photorealistic person. Moreover, we provide for significantly higher resolution results ($512\times512$ vs. a resolution of $176\times256$ or $64\times128$ in the leading pose transfer benchmarks), over images with substantial pose variation.

The application domain we focus on is the insertion of a target person into an image that contains other people. This is a challenging application domain, since it is easy to spot discrepancies between the novel person in the generated image, and the existing ones. In contrast, methods that generate images from scratch enjoy the ability to generate ``convenient images''. 

In addition, images that are entirely synthetic are judged less harshly, since the entire image has the same quality. In our case, the generated pixels are inserted into an existing image and can therefore stand out as being subpar with respect to the quality of the original image parts. Unlike other applications, such as face swapping, our work is far less limited in the class of objects. 

Similar to face swapping and other guided image manipulation techniques, the appearance of the output image is controlled by that of an example.  However, the appearance in our case is controlled by multiple components: the face, several clothing items, and hair. 

Our method employs three networks. The first generates the pose of the novel person in the existing image, based on contextual cues that pertain to the other persons in the image. The second network renders the pixels of the new person, as well as a blending mask. Lastly, the third augments the face of the target person in the generated image in order to ensure artifact-free faces.

In an extensive set of experiments, we demonstrate that the first of our networks can create poses that are indistinguishable from real poses, despite the need to take into account the social interactions in the scene. The first and second networks provide a state of the art solution for the pose transfer task, and the three networks combined are able to provide convincing ``wish you were here'' results, in which a target person is added to an existing photograph.

The method is trained in an unsupervised manner, in the sense that unlike previous work, such as networks trained on the DeepFashion dataset, it trains on single images, which do not present the same person in different poses. However, the method does employ a set of pretrained networks, which were trained in a fully supervised way, to perform various face and pose related tasks: a human body part parser, a face keypoint detector, and a face-recognition network.

Our main contributions are: (i) the first method, as far as we can ascertain, is the first to generate a human figure in the context of the other persons in the image, (ii) a person generating module that renders a high resolution image and mask, given two types of conditioning, the first being the desired multi-labeled shape in the target image, and the second being various appearance components, (iii) the ability to perform training on a set of unlabeled images ``in the wild'', without any access to paired source and target images, by utilizing existing modules trained for specific tasks, (iv) unlike recent pose transfer work, which address a simpler task, we work with high resolution images, generating $512\times512$ images, (v) our results are demonstrated in a domain in which the pose, scale, viewpoint, and severe occlusion vary much more than in the pose transfer work from the literature, and (vi) demonstrating photo realistic results in a challenging and so far unexplored application domain.

Our research can be used to enable natural remote events and social interaction across locations. AR applications can also benefit from the addition of actors in context. Lastly, the exact modeling of relationships in the scene can help recognize manipulated media.

\section{Related work}

There is considerably more work on the synthesis of novel images, than on augmenting existing views. A prominent line of work generates images of human subjects in different poses~\cite{balakrishnan2018synthesizing, human_dynamics18}, which can be conditioned on a specific pose~\cite{chan2018dance, guided, li2019dense}. The second network we employ (out of the three mentioned above) is able to perform this task, and we empirically compare with such methods. 
Much of the literature presents results on the DeepFashion dataset~\cite{liuLQWTcvpr16DeepFashion}, in which a white background is used. In the application we consider, it is important to be able to smoothly integrate with a complex scene. However, for research purposes only and for comparing with the results of previous work~\cite{ma2017pose,siarohin2018deformable,esser2018variational,zhu2019progressive,dong2018soft}, we employ this dataset.

Contributions that include both a human figure and a background scene, include  vid2vid~\cite{wang2018vid2vid} and the "everybody dance now" work~\cite{chan2018dance}. These methods learn to map between a driver video and an output video, based on pose or on facial motion. Unlike the analog pose-to-image generation part of our work, in~\cite{wang2018vid2vid,chan2018dance} the reference pose is extracted from a real frame, and the methods are not challenged with generated poses. Our method deals with generated poses, which suffer from an additional burden of artifacts. In addition, the motion-transfer work generates an entire image, which includes both the character and the background, resulting in artifacts near the edges of the generated pose~\cite{arbitrary, posetransfer}, and the loss of details from the background. In our work, the generated figure is integrated with the background using a generated alpha-mask.


Novel generation of a target person based on a guiding pose was demonstrated by Esser et al., who presented two methods for mixing the appearance of a figure seen in an image with an arbitrary pose~\cite{esser2018variational,human3d}. Their methods result in a low-resolution output with noticeable artifacts, while we work at a higher resolution of 512p.  The work of Balakrishanan et al. also provides lower resolution outputs, which are set in a specific background~\cite{balakrishnan2018synthesizing}. In our experiment, we compared against the recent pose transfer work~\cite{ma2017pose,siarohin2018deformable,zhu2019progressive}.

A semantic map based method for human generation was presented by~\cite{dong2018soft}. Contrary to our method, this work was demonstrated solely on the lower resolution, and lower pose variation datasets of DeepFashion and Market-1501 ($176\times256$ and $64\times128$). Additionally, the target encoding method in~\cite{dong2018soft} relies on an additional semantic map, identical to the desired target person, requiring the target person to be of the same shape, which precludes other applications, such as component replacement. Moreover, the previous method requires the pose keypoints, which increases the complexity of the algorithm, and limits the application scope, such as the one that we show for drawing a person.

As far as we know, no literature method generates a human pose in the context of other humans in the scene.

\section{Method}
Given a source image $x$, the full method objective is to embed an additional person into the image, such that the new person is both realistic, and coherent in context. The system optionally receives a coarse position for the new person, in the form of a bounding box $b$. This allows for crude control over the new person position and size, yet still leaves most of the positioning for the algorithm. 

We employ three phases of generation, in which the inserted person becomes increasingly detailed. The Essence Generation Network (EGN) generates the semantic pose information of the target person $p$ in the new image, capturing the scene essence, in terms of human interaction. The Multi-Conditioning Rendering Network (MCRN) renders a realistic person, given the semantic pose map $p$, and a segmented target person, which is given as a multi-channel tensor $t$. The Face Refinement Network (FRN) is used to refine the high-level features of the generated face $f$, which requires special attention, due to the emphasis given to faces in human perception.

\begin{figure*}
  \centering
 \includegraphics[width=0.86\textwidth]{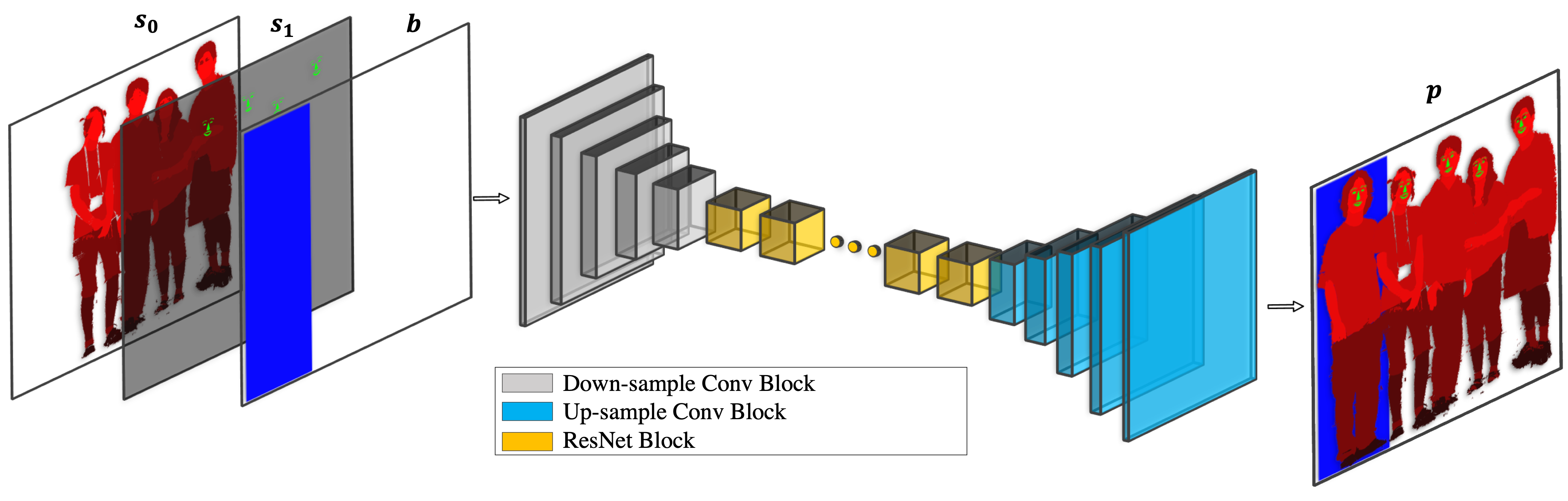}
  \caption{The architecture of the Essence Generation Network. Given a body and face semantic maps $s$, and an optional bounding-box $b$, the network generates the semantic map $p$ of a novel person, which is correlated in context to the human interaction in the scene. The generated person is highlighted in blue.}
  \label{fig:arch_p2p}
\end{figure*}

\subsection{Essence generation network}

The Essence Generation Network (EGN) is trained to capture the human interaction in the image, and generate a coherent way for a new human to join the image. Given a two-channel semantic map $s$ of the input image $x$ with a varying number of persons, and an optional binary third channel $b$ containing a bounding box, the network generates the two-channel semantic map of a new person $p$, which is compatible with the context of the existing persons, as seen in Fig.~\ref{fig:arch_p2p},~\ref{fig:wish_p2p_nobbx_results}.

More precisely: both $s$ and $p$ contain one channel for the person's semantic map, and one face channel, derived from facial keypoints. $s$ pertains to the one or more persons in the input image, while $p$ refers to the novel person. The semantic map, i.e., the first channel of $s$ and $p$, is reduced to eight label groups, encoded as the values $0,36,72,..,252$. These represent the background (0), hair, face, torso and upper limbs, upper-body wear, lower-body wear, lower limbs, and finally shoes. The choice of this reduced number of groups is used to simplify semantic generation, while still supporting detailed image generation. 

The face channel of $s$ and $p$ is extracted by considering the convex hulls over the detected facial keypoints, obtained by the method of~\cite{cao2018openpose}. The third channel $b$ is optional, and contains a bounding box, indicating the approximate size and position of the new person in $p$. During training, the bounding box is taken as the minimal and maximal positions of the labels in the x and y axes. Both the face and bounding-box channels are binary and have values that are either $0$ or $255$.

We train two EGN {\color{black} models ($EGN$ and $EGN'$) in parallel} to perform the following mapping:
\begin{equation}
    p=EGN(s,f,b)\quad \text{~or~} \quad p=EGN'(s,f)
\end{equation} 
where $EGN$ obtains one additional input channel in comparison to $EGN'$. For brevity, we address $EGN$ below. The input tensors are resized to the spatial dimensions of $368\times 368$ pixels. The subsequent networks employ higher resolutions, generating high resolution images. The EGN encoder-decoder architecture is based on the one of pix2pixHD~\cite{wang2018pix2pixHD} with two {\color{black}major} modifications. First, the VGG feature-matching loss is disabled, as there is an uncertainty of the generated person. In other words, given a source image, there is a large number of conceivable options for a new person to be generated, in the context of the other persons in the scene. These relations are captured by the discriminator loss as well as the discriminator feature-matching loss, as both losses receive both $(s,f,b)$ and $p$. The second modification is the addition of a derivative regularization loss $\mathcal{L}^{p}_{\nabla}=\|p_x\|_1 + \|p_y\|_1$, which is applied over the first channel of $p$. This loss minimizes the high-frequency patterns in the generated semantic map image.

\subsection{Multi-conditioning rendering network}
\begin{figure*}
  \centering
  \begin{tabular}{c|c}
 \includegraphics[width=0.605\textwidth]{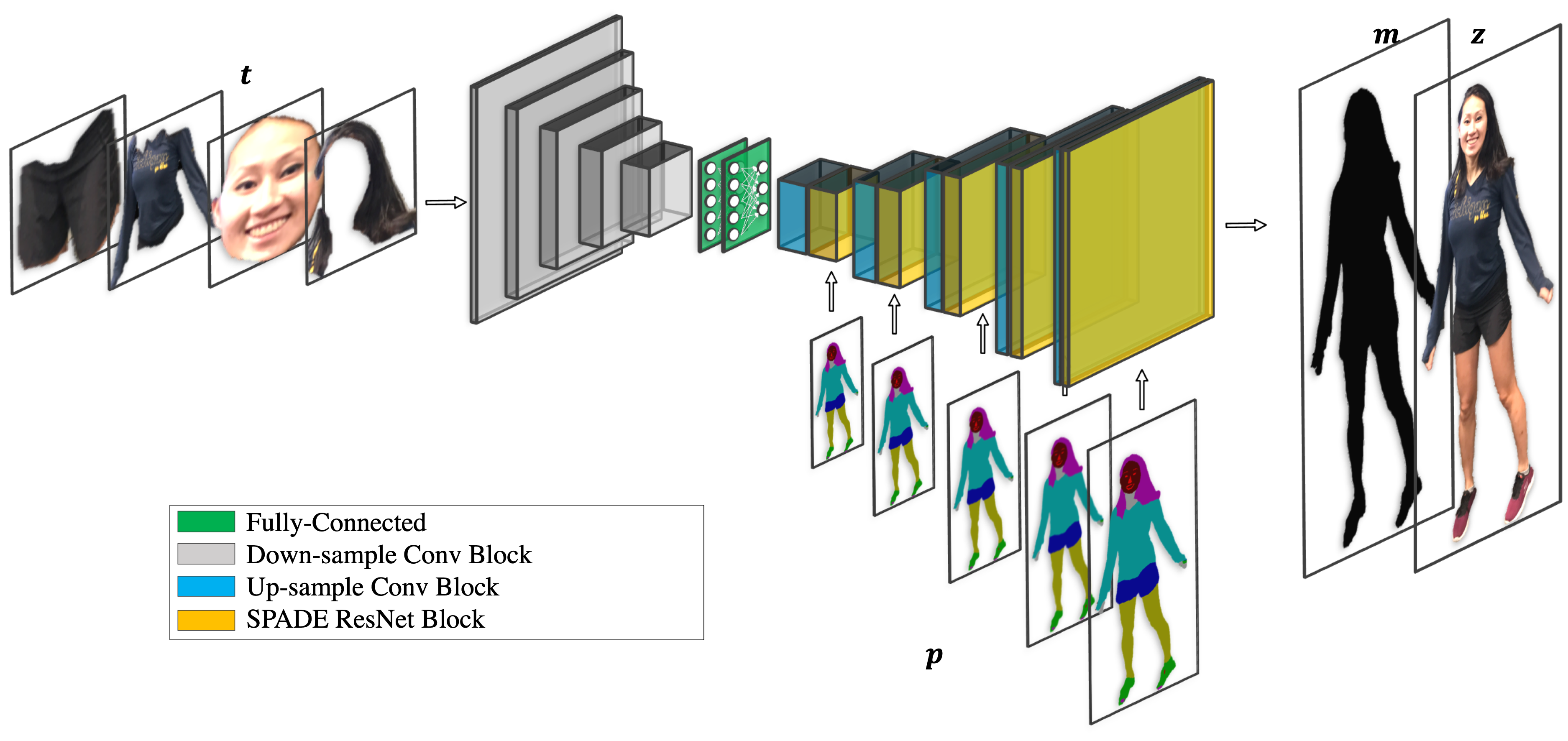}&
 \includegraphics[width=.304\textwidth]{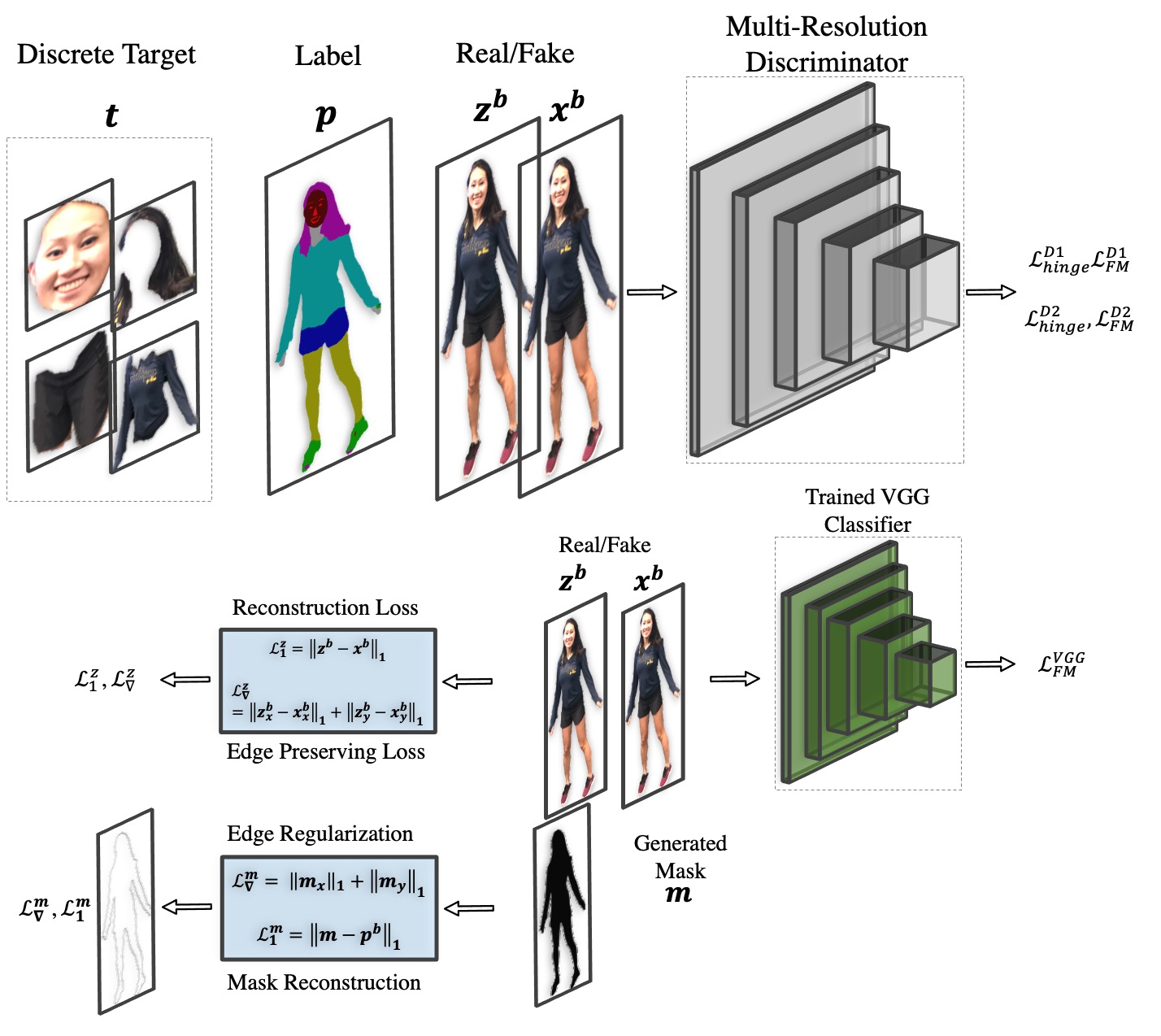} \\
 (a) & (b)\\
 \end{tabular}
  \caption{(a) MCRN's architecture. Given an input target $t$, and a conditioning semantic map $p$, a person $z$ and blending mask $m$ are rendered. The mask $m$ is then employed to blend the rendered person $z$ into a final image $o$. (b) The loss terms used to train MCRN.}
  \label{fig:arch_p2f}
\end{figure*}

The MCRN mapping is trained to render and blend a realistic person into the input image $x$, creating a high-resolution ($512\times512$) image $o$.  It is given a conditioning signal in the form of a semantic pose map $p$, and an input specifying the parts of a segmented person $t$, see Fig.~\ref{fig:arch_p2f}(a). The conditioning signal $p$, which is generated by the EGN at inference time, is introduced to the decoder part of MCRN through SPADE blocks~\cite{park2019SPADE}. This conditioning signal acts as the structural foundation for the rendered person image $z$, and the corresponding mask $m$. 

The segmented person $t$ is incorporated through the MCRN encoder, which embeds the target person appearance attributes into a latent space. $t$ allows for both substantial control over the rendered person $z$ (e.g. replacing the person's hair or clothing, as seen in Fig.~\ref{fig:replace_short_hair_shirt_pants_df1}, and supplementary Fig.~{\color{red}1,2,3}). The segmented structure of $t$ has the advantage over simply passing the image of the target person, in that it does not allow for a simple duplication of the target person in the output image $z$. This property is important, as during training we employ the same person in both the target output image, and as the input to MCRN. 

The tensor $t$ is of size $6\times 3\times 128\times 128$, which corresponds to the six semantic segmentation classes (hair, face, upper-body wear, lower-body wear, skin, and shoes), three RGB channels each, and a spatial extent of $128^2$ pixels. Each of the six parts is obtained by cropping the body part using a minimal bounding box, and resizing the crop to these spatial dimensions. 

To preempt a crude insertion of the generated person $z$ into the image output $o$ and avoid a "pasting" effect, the network generates a learnable mask $m$ in tandem with the rendered image of the person $z$. The output image is therefore generated as:
\begin{equation}
    [z,m] = MCRN(t,p),\quad o=x\odot (1-m) + z\odot m
\end{equation}

The mask $m$ is optimized to be similar to the binary version of the pose image $p$, which is denoted by $p^b$. For this purpose, the L1 loss is used
$\mathcal{L}^{m}_1=\|m - p^b\|_1$. Additionally, the mask is encouraged to be smooth as captured by the loss 
\begin{equation}
    \mathcal{L}^{m}_{\nabla}=\|m_x\|_1 + \|m_y\|_1.
    \label{eq:lnabla}
\end{equation} 

The architecture of the MCRN encoder is composed of five consecutive (Conv2D,InstanceNorm2D~\cite{ulyanov2016instance}) layers, followed by an FC layer with a LeakyReLU activation, resulting in a latent space the size of $256$. The latent space is processed through an additional FC layer, reshaped to a size of $4x4x1024$. The decoder has seven upsample layers with interleaving SPADE blocks. It is trained using the loss terms depicted in Fig.~\ref{fig:arch_p2f}(b). Namely:
\begin{equation}
\mathcal{L}^{G}_{hinge}=-\|D_{1,2}(t,p,z^b)\|_1 
\end{equation}
\begin{equation}
\begin{split}
\mathcal{L}^{D_{1,2}}_{hinge}=-\|\min(D_{1,2}(t,p,z^b)-1, 0)\|_1 - \\ \|\min(-D_{1,2}(t,p,x^b)-1, 0)\|_1  
\end{split}
\end{equation}
\begin{equation}
\mathcal{L}^{D_{k=1,2}}_{FM}=\mathbb{E}_{(t,p,x^b,z^b)}\sum_{j=1}^{M}\frac{1}{N_j}||D_k^{(j)}(t,p,x^b)-  D_k^{(j)}(t,p,z^b))||_1
\end{equation}
with $M$ being the number of layers, $N_j$ the number of elements in each layer, $D_k^{(j)}$ the activations of discriminator $k$ in layer $j$, and $z^b,x^b=z \odot p^b,x \odot p^b$.
\begin{multline}
\mathcal{L}^{VGG}_{FM}=\sum_{j=1}^{M}\frac{1}{N'_j}||VGG^{(j)}(x)-VGG^{(j)}(o))||_1
\end{multline}
with $N'_j$ being the number of elements in the $j$-th layer, and $VGG^{(j)}$ the VGG classifier activations at the $j$-th layer.
\begin{multline}
\mathcal{L}^{z}_{1}=\|z^b-x^b\|_1,
\mathcal{L}^{z}_{\nabla}=\|z^b_x-x^b_x\|_1+\|z^b_y-x^b_y\|_1 
\end{multline}

\subsection{Face refinement network}

The third network, FRN, receives as input the face crop of the novel person in $o$, as well as a conditioning signal that is the face descriptor of the target face, as obtained from the original image of the target person $y$ (before it was transformed to the tensor $t$). For that purpose, the pretrained VGGFace2~\cite{vggface2} network is used, and the activations of the penultimate layer are concatenated to the FRN latent space.

FRN {\color{black}applies} the architecture of~\cite{de-id}, which employs the same two conditioning signals, for a completely different goal. While in~\cite{de-id}, the top level perceptual features of the generated face $f$, obtained from the embedding $e$ of the VGGFace2 network, are distanced from those of the face $f_y$ in $y$, in our case, the perceptual loss encourages the two to be similar by minimizing the distance $\|e(f)-e(f_y)\|_1$.

FRN's output is blended with a second mask $m^f$ as:
\begin{equation}
    \begin{split}
        [f,m^f] = FRN(c(o),VGGFace2(c(y)))\\
        w = o \odot (1-m^f) + f \odot(m^f)
    \end{split}
\end{equation}
where $c$ is the operator that crops the face bounding box.

\section{Experiments}

\begin{figure*}
  \centering
 \includegraphics[width=0.9\textwidth]{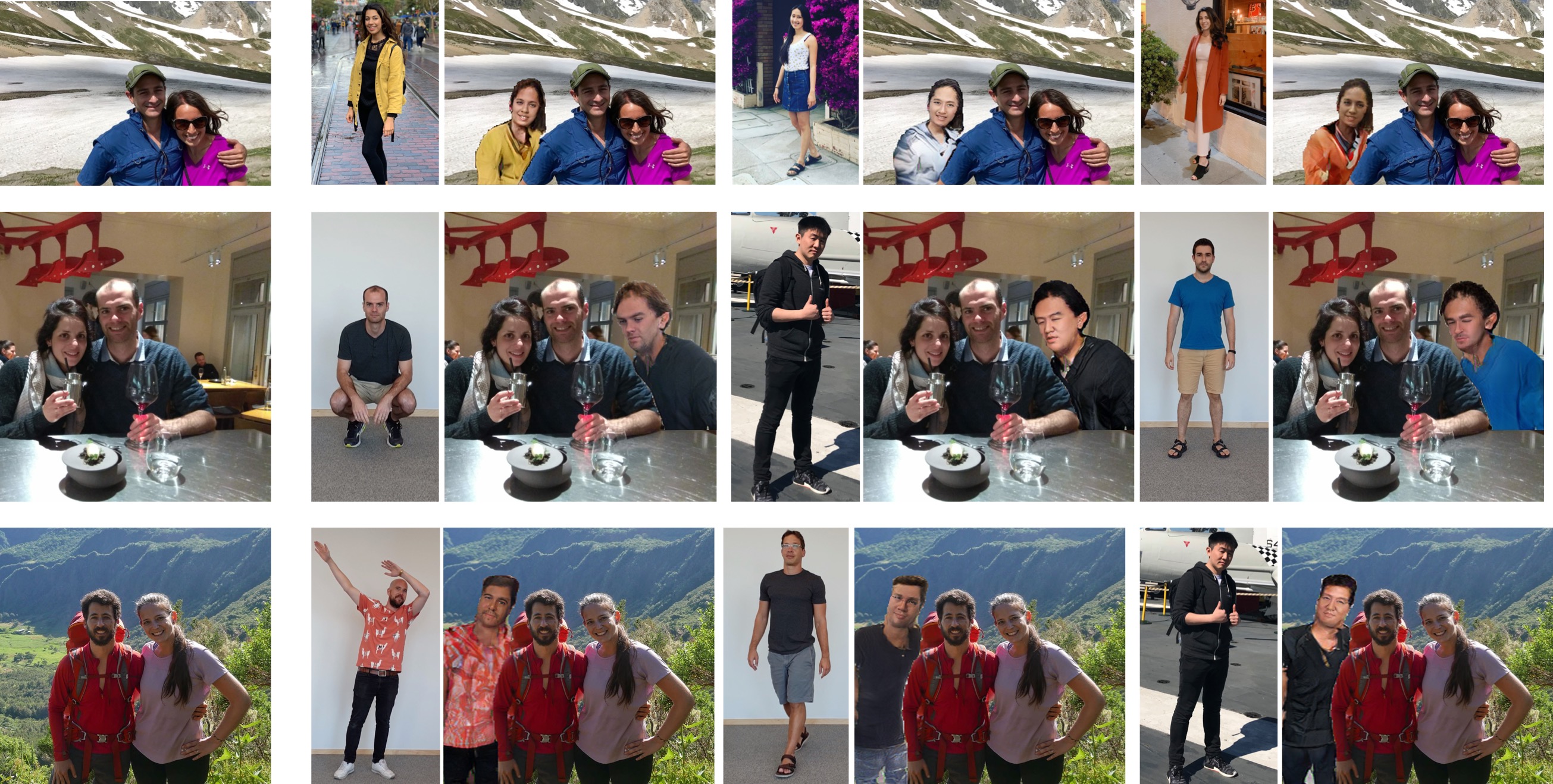} 
  \caption{"Wish you were here" samples. Each shows has a source image $x$, and 3 different pairs of inserted person $y$ and output image $w$.}
  \label{fig:wish_results_men}
\end{figure*}

\begin{figure*}
  \centering
 \includegraphics[width=0.9\textwidth]{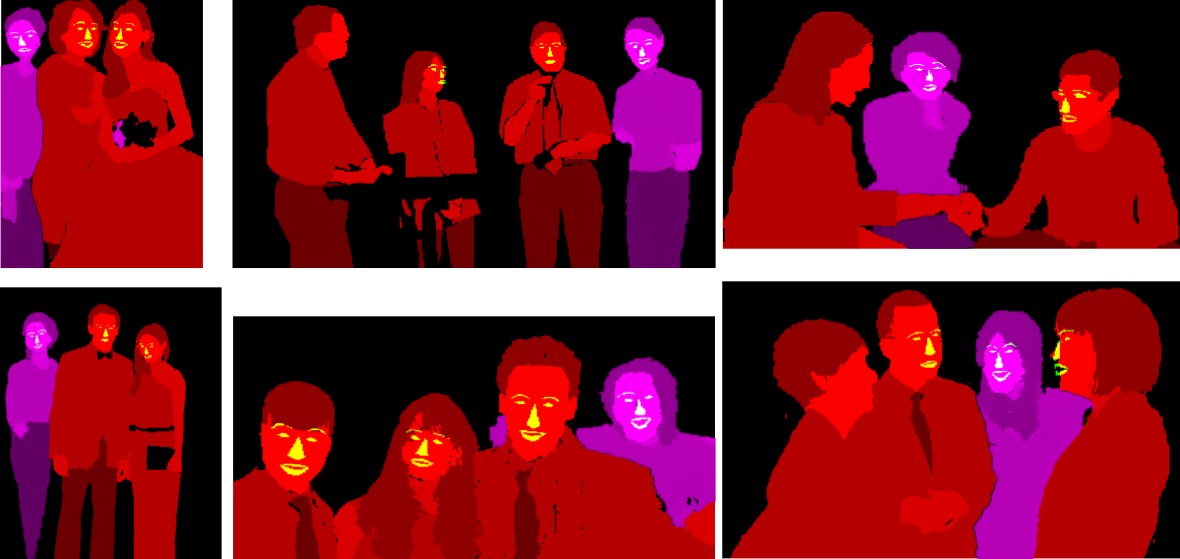} 
  \caption{Unconstrained (no bounding-box) samples of EGN'. For each input (red), the generated pose (purple) is shown.}
  \label{fig:wish_p2p_nobbx_results}
\end{figure*}

\begin{figure}
  \centering
 \includegraphics[width=.9\linewidth]{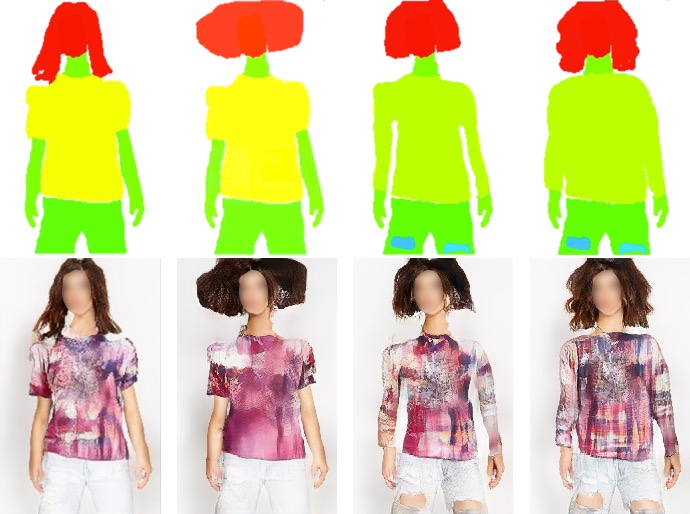} 
  \caption{Drawing a person (DeepFashion). A semantic map is crudely drawn (row 1) utilizing the annotation tool of~\cite{zhao2018understanding}, distinguishing between the hair (orange), face (red), torso/upper-limbs (bright-green 1), T-shirt (yellow), sweat-shirt (bright-green 2), pants (green), lower-limbs (blue). The rendered person generated by the MCRN (row 2) conforms to the conditioning segmentation, despite the deviation from the original dataset. The facial keypoints (not shown here) are taken from a randomly detected image. A video depicting the drawing and generation process is attached in the supplementary.}
  \label{fig:draw_person_df}
\end{figure}

\begin{figure*}
  \centering
 \includegraphics[width=0.7\textwidth]{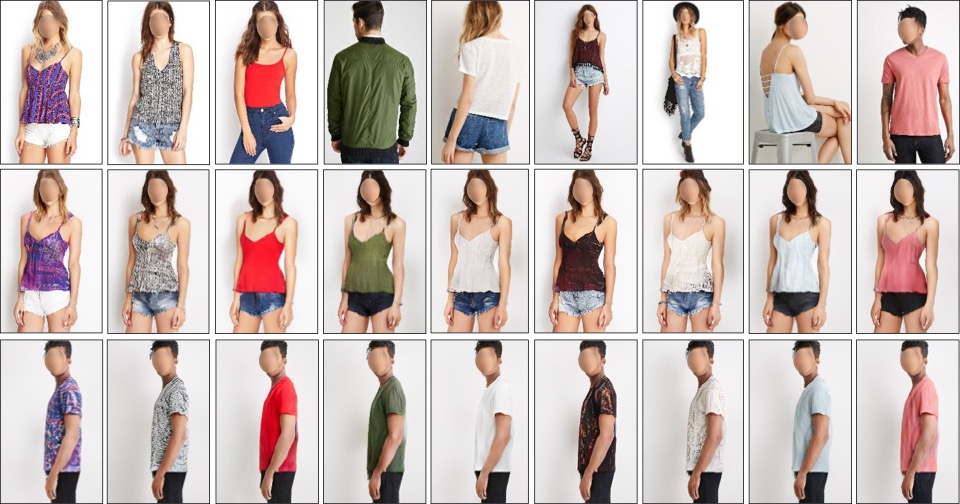} 
  \caption{Replacing the hair, shirt, and pants (DeepFashion). For each target $y$ (row 1), the hair, shirt and pants (row 2), shirt only (row 3), are replaced for the semantic map $s$ of the upper-left and upper-right person. EGN/FRN are not used. See also Fig. 2,3,4 in the supp.}
  \label{fig:replace_short_hair_shirt_pants_df1}
\end{figure*}


\begin{figure*}[t]
  \centering
  \begin{tabular}{@{}c@{~}c@{~}c@{~}c@{~}c@{~}c@{~}c@{}}
  \includegraphics[width=1.2654316936cm]{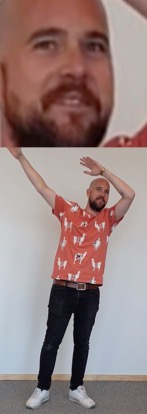} & \includegraphics[width=2.2cm]{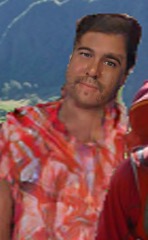} & \includegraphics[width=2.2cm]{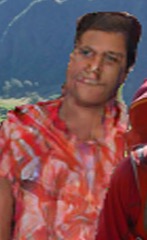} & \includegraphics[width=2.2cm]{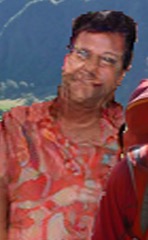} & \includegraphics[width=2.2cm]{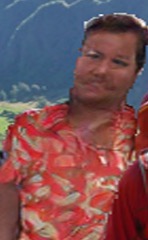} & 
  \includegraphics[width=2.2cm]{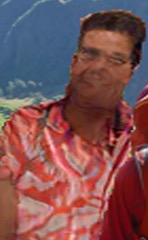} & \includegraphics[width=2.2cm]{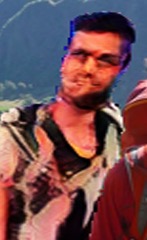} \\
  (a) & (b) & (c) & (d) & (e) & (f) & (g) 
  \end{tabular} 
  \caption{MCRN ablation study. (a) Target person, (b) our result, (c) no FRN (\textit{distorted face, does not resemble target}), (d) no $L^z_{1}$ and $L^z_{\nabla}$ (\textit{blurry face, distorted skin patterns}), (e) $L^m_{\nabla}$ not tuned (\textit{strong edges pixelization}), (f) no mask (\textit{unnatural blending ``pasting" effect''}), (g) no segmented encoder (\textit{excessive artifacts stemming from target and label spatial difference}).}
  \label{fig:p2f_ablation}
\medskip
  \centering
  \begin{tabular}{@{}c@{~}c@{~}c@{~}c@{~}c@{~}c@{~}c@{}}
  \includegraphics[width=2.2cm]{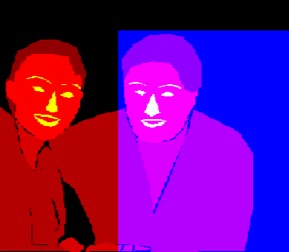} & \includegraphics[width=2.2cm]{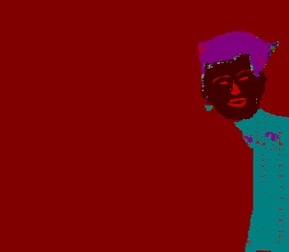} & \includegraphics[width=2.2cm]{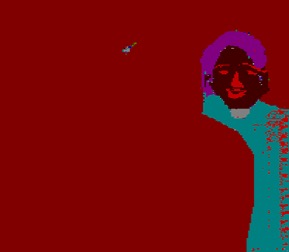} &
  \includegraphics[width=2.2cm]{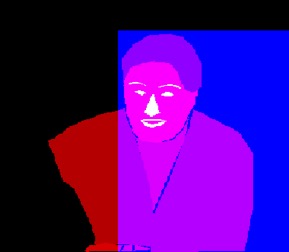} & 
  \includegraphics[width=2.2cm]{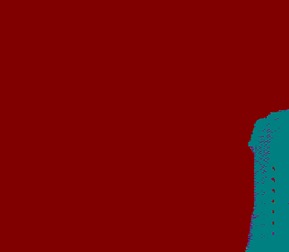} & 
  \includegraphics[width=2.2cm]{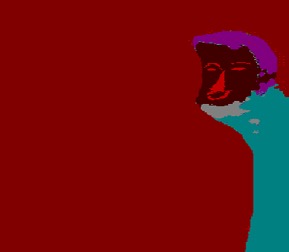} & \includegraphics[width=2.2cm]{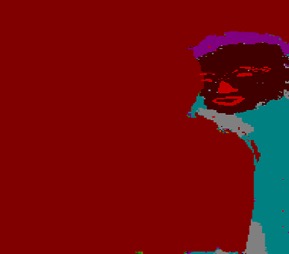} \\
  (a) & (b) & (c) & (d) & (e) & (f) & (g) 
  \end{tabular} 
  \caption{EGN ablation study. (a) Semantic map input for (b)-(c), (b) our result, (c) no $L^s_{\nabla}$ (\textit{high-frequency patterns, as well as isolated objects generated}), (d) semantic map input for (e)-(g), (e) single person input (\textit{context can be less descriptive}), (f) VGG feature-matching enabled (\textit{shape is matched regardless of deformation artifacts}), (g) generation shape reduced to $256\times 256$ (\textit{labels are perforated, new labels generated on top of existing segmentations}). Columns (b)-(c) and (e)-(g) are presented in high-contrast colors for clarity.}
  \label{fig:p2p_ablation}
\end{figure*}

Both the EGN and MCRN are trained on the Multi-Human Parsing dataset (\cite{li2017towards},~\cite{zhao2018understanding}). We choose this as our primary dataset, due to the high-resolution images and the diverse settings, in terms of pose, scene, ethnicity, and age, which makes it suitable for our task. We randomly select $20,000$ images for training, and $250$ images for testing. 
EGN is trained such that for each sample, all semantic maps in $s$ are maintained, excluding one, which is the generation objective $p$. In addition, we filter-out images that do not contain at least one detected set of facial keypoints. Overall, we obtain $51,717$ training samples,  training for 300 epochs, with a batch size of 64. MCRN is trained on each person separately, resulting in $53,598$ sampled images. The network is trained for 200 epochs, with a batch size of 32.

Our method has a single tuning parameter. This is the strength of the mask edge regularization (Eq.~\ref{eq:lnabla}) . The scale of the loss term was set during the development process to be multiplied by a factor of $5$, after being tested with the values of $[0, 1, 5, 10]$. This value is verified in the MCRN ablation study in Fig.~\ref{fig:p2f_ablation}.

\noindent{\bf Context-aware generation.} We provide samples for a variety of target persons $y$ in the full context-aware generation task, in Fig.~\ref{fig:teaser},~\ref{fig:wish_results_men}. In these experiments, EGN is given a random bounding-box $b$, with a size and y-axis location randomly selected to be between $0.9$ to $1.1$ of an existing person in the image, while the x-axis location is randomly selected by a uniform distribution across the image. The EGN generates a semantic map $p$, which is then run by the MCRN for various targets $y$, shown for each column. FRN is then applied to refine the rendered face. As can be observed by the generated results, EGN felicitously captures the scene context, generating a semantic map of a new person that is well-correlated with the human interactions in the scene. MCRN successfully renders a realistic person, as conditioned by the target $y$, and blends the novel person well, as demonstrated over diverse targets.



The case with no specified input bounding box is demonstrated in Fig.~\ref{fig:wish_p2p_nobbx_results}. As can be observed, EGN' selects highly relevant poses by itself.

\noindent{\bf Individual component replacement.} Evaluating the MCRN ability to generalize to other tasks, we utilize it for hair, shirt, and pants replacement, demonstrated over both the DeepFashion dataset~\cite{liuLQWTcvpr16DeepFashion} in Fig.~\ref{fig:replace_short_hair_shirt_pants_df1}, supplementary Fig.~{\color{red}2},~{\color{red}3}, and high-resolution in supplementary Fig.~{\color{red}4}.
As seen in the latter dataset, MCRN can be successfully applied to unconstrained images, rather than low-variation datasets only, such as DeepFashion, increasing the applicability and robustness of this task. We employ the model of~\cite{liang2018look,Gong_2017_CVPR} for human parsing.

\noindent{\bf Person drawing.}
An additional application of the MCRN is free-form drawing of a person. We intentionally demonstrate this task over a set of extreme, and crudely drawn sketches, depicting the ability to render persons outside of the dataset manifold, yet resulting in coherent results, as seen in Fig.~\ref{fig:draw_person_df}, and the supplementary video. The annotation tool presented in~\cite{zhao2018understanding} is used to sketch the semantic map, and a video depicting the drawing and generation process is attached as supplementary.

\noindent{\bf Pose transfer evaluation.} 
MCRN can be applied to the pose transfer task. By modifying EGN to accept as input the concatenation of a source semantic map, source pose keypoints (a stick figure, as extracted by the method of~\cite{cao2018openpose}), and target pose keypoints, we can generate the target semantic map $p$, which is then fed into MCRN. A DensePose~\cite{Guler2018DensePose} representation can be used instead of the stick-figure as well. 

A qualitative comparison of this pipeline to the methods of~\cite{zhu2019progressive,ma2017pose,esser2018variational,siarohin2018deformable} is presented in supplementary Fig.~{\color{red}4}. The work of~\cite{dong2018soft} presents visually compelling results, similar to ours in this task. We do not present a qualitative comparison to~\cite{dong2018soft} due to code unavailability. However, a quantitative comparison is presented in Tab.~\ref{tab:deepfashion_comp} (FRN is not applied). 

Providing reliable quantitative metrics for generation tasks is well known to be challenging. Widely used methods such as Inception Score~\cite{gantricks} and SSIM~\cite{wang2004image} do not capture perceptual notion, or human-structure~\cite{barratt2018note,zhu2019progressive}. Metrics capturing human-structure such as PCK~\cite{yang2012articulated}, or PCKh~\cite{andriluka20142d} have been proposed. However, they rely on a degenerated form of the human form (keypoints). 

We therefore develop two new dense-pose based human-structure metrics (DPBS and DPIS), and provide the Python code in the supplementary. Additionally, we evaluate perceptual notions using the LPIPS (Learned Perceptual Image Patch Similarity) metric~\cite{zhang2018perceptual}. DPBS (DensePose Binary Similarity) provides a coarse metric between the detected DensePose~\cite{Guler2018DensePose} representation of the generated and ground-truth images, by computing the Intersection over Union (IoU) of the binary detections. The second novel metric, DPIS (DensePose Index Similarity), provides a finer shape-consistency metric, calculating the IoU of body-part indices, as provided by the DensePose detection. The results are then averaged across the bodyparts.

The quantitative comparison follows the method described by~\cite{zhu2019progressive} in terms of dataset split into train and test pairs (101,966 pairs are randomly selected for training and 8,570 pairs for testing, with no identity overlap between train and test). Our method achieves the best results in terms of perceptual metrics out of the tested methods (both for our keypoint and DensePose based methods). For human-structural consistency, both our methods achieve top results for the DPBS metric, and highest for the DensePose based model in the DPIS metric. Our methods scores well for the controversial metrics (SSIM, IS) as well.

\begin{table}
\begin{small}
\centering
  \begin{tabular}{l@{~}c@{~}c@{~}c@{~}c@{~}c@{~}c}
    \toprule
    
    Method & $\downarrow$ LPIPS & $\downarrow$ LPIPS & $\uparrow$ DPBS & $\uparrow$ DPIS & $\uparrow$ SSIM & $\uparrow$ IS \\
    &  (SqzNet) &  (VGG)   \\
    \midrule
    Ma~\cite{ma2017pose} & 0.416 & 0.523  & 0.791  & 0.392  & 0.773  & 3.163 \\
    Siarohin~\cite{siarohin2018deformable} & - & - & - & - & 0.760 & 3.362 \\
    Esser~\cite{esser2018variational} &  - & - & - & - & 0.763 & \textbf{3.440} \\
    Zhu~\cite{zhu2019progressive} &  0.170  & 0.299 & 0.840 & \textit{0.463} & 0.773 & 3.209 \\
    Dong~\cite{dong2018soft} &  - & - & - & - & \textbf{0.793} & 3.314 \\
    Ours (DP) &  \textbf{0.149} & \textbf{0.264} & \textbf{0.862} & \textbf{0.470} & \textbf{0.793} & \textit{3.346} \\
    Ours (KP) &  \textit{0.156} & \textit{0.271} & \textit{0.852} & 0.448 & \textit{0.788} & 3.189 \\
    \midrule
    
  \end{tabular}
  
    \end{small}
\hfill
\caption{Pose-transfer on the DeepFashion dataset. Shown are the LPIPS~\cite{zhang2018perceptual}, DPBS, DPIS, SSIM~\cite{wang2004image}, and IS~\cite{gantricks} metrics. Both our DensePose (DP) and keypoint (KP) based methods achieve state-of-the-art results in most metrics. FRN is not applied.}
  \label{tab:deepfashion_comp}
\end{table}


\begin{table}
\begin{tabular}{@{}c@{~~~~~~}c@{}}
\begin{tabular}[b]{@{}c@{~~}c@{}}

    \toprule
    $N$     & Success \\
    \midrule
        
3 &	$39.47\% \pm 0.47$ \\
4 &	$47.37\% \pm 0.49$ \\
5 &	$28.07\% \pm 0.43$ \\
6 &	$47.37\% \pm 0.45$ \\

    \midrule
Average    & $42.98\% \pm 0.47$ \\ 
    \bottomrule
  \end{tabular} 
&
\includegraphics[width=0.5\linewidth,height=3.1cm]{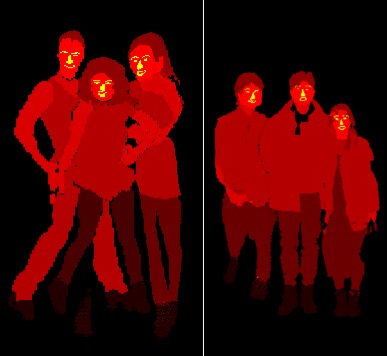}
\\
  (a) & (b)\\
  \end{tabular}
  \caption{User study. (a) Success rate in user recognition of the generated person. Shown per $N$ number of persons in an image. (b) Examples of images used. For each image, the user is given unlimited time to identify the generated person.}
    \label{tab:user_study}

\end{table}

\noindent{\bf User Study.}
A user study is shown in Tab.~\ref{tab:user_study}, presented per $N=3,4,5,6$ number of persons in an image (including the generated person). For each image, the user selects the generated person. The user is aware that all images contain a single generated person, and contrary to user studies commonly used for image generation, no time constraints are given. The low success rate validates EGN's ability to generate novel persons in context. Note that the success rate does not {\color{black}correlate} with $N$ as expected, perhaps since the scene becomes more challenging to modify the larger $N$ is.

\noindent{\bf Ablation study} 
We provide qualitative ablation studies for both EGN and MCRN. As the "wish your were here" application does not have a ground-truth, perceptual comparisons, or shape-consistency quantitative methods do not capture the visual importance of each component. Other methods that do not rely on a ground-truth image (e.g. Inception Score, FID), 
are unreliable, as for the pose-transfer task, higher IS seems correlated with more substantial artifacts, indicating that a higher degree of artifacts results in a stronger perceived diversity by the IS.

The MCRN ablation is give in Fig.~\ref{fig:p2f_ablation}, showcasing the importance of each component or setting. Details are given in the figure caption.

The EGN ablation is given in Fig.~\ref{fig:p2p_ablation}. For the generated person, there are numerous generation options that could be considered applicable in terms of context. This produces an involved ablation study, encompassing additional deviations between tested models, that are not a direct result of the different component tested. Observing beyond the minor differences, the expected deviations (as seen throughout the experiments performed to achieve the final network) are detailed in the figure caption.


\section{Discussion}
Our method is trained on cropped human figures. The generated figure tends to be occluded by other persons in the scene, and does not occlude them. The reason is that during training, the held-out person can be occluded, in which case the foreground person(s) are complete. {\color{black} Alternatively,} the held-out person can be complete, in which case, once removed, the occluded person(s) appear to have missing parts. At test time, the persons {\color{black}contain missing areas that are solely due to the existing scene}. Therefore, test images appear as images in which the held-out person is occluded.

In a sense, this is exactly what the ``wish you were here'' application (adding a person to an existing figure) calls for -- finding a way to add a person, without disturbing the persons already there. However, having control over the order of the persons in the scene relative to the camera plane, would add another dimension of variability.


A limitation of the current method, is that the generated semantic map $p$ is not conditioned on the target person $y$ or their attributes $t$. Therefore, for example, the hair of the generated figure is not in the same style as the target person. This limitation is not an inherent limitation, as one can condition EGN on more inputs, but rather a limitation of the way training is done. Since during training we have only one image, providing additional appearance information might impair the network generalization capability.  A partial solution may be to condition, for example, on very crude descriptors such as the relative hair length. 

\section{Conclusions}

We demonstrate a convincing ability to add a target person to an existing image. The method employs three networks that are applied sequentially, and progress the image generation process from the semantics to the concrete.

From a general perspective, we demonstrate the ability to modify images, adhering to the semantics of the scene, while preserving the overall image quality.

{\small
\bibliographystyle{ieee_fullname}
\bibliography{gans}
}

\clearpage


\onecolumn

\appendix




\section{Additional individual component replacement samples}

\begin{figure}[H]
  \centering
 \includegraphics[width=0.9\textwidth]{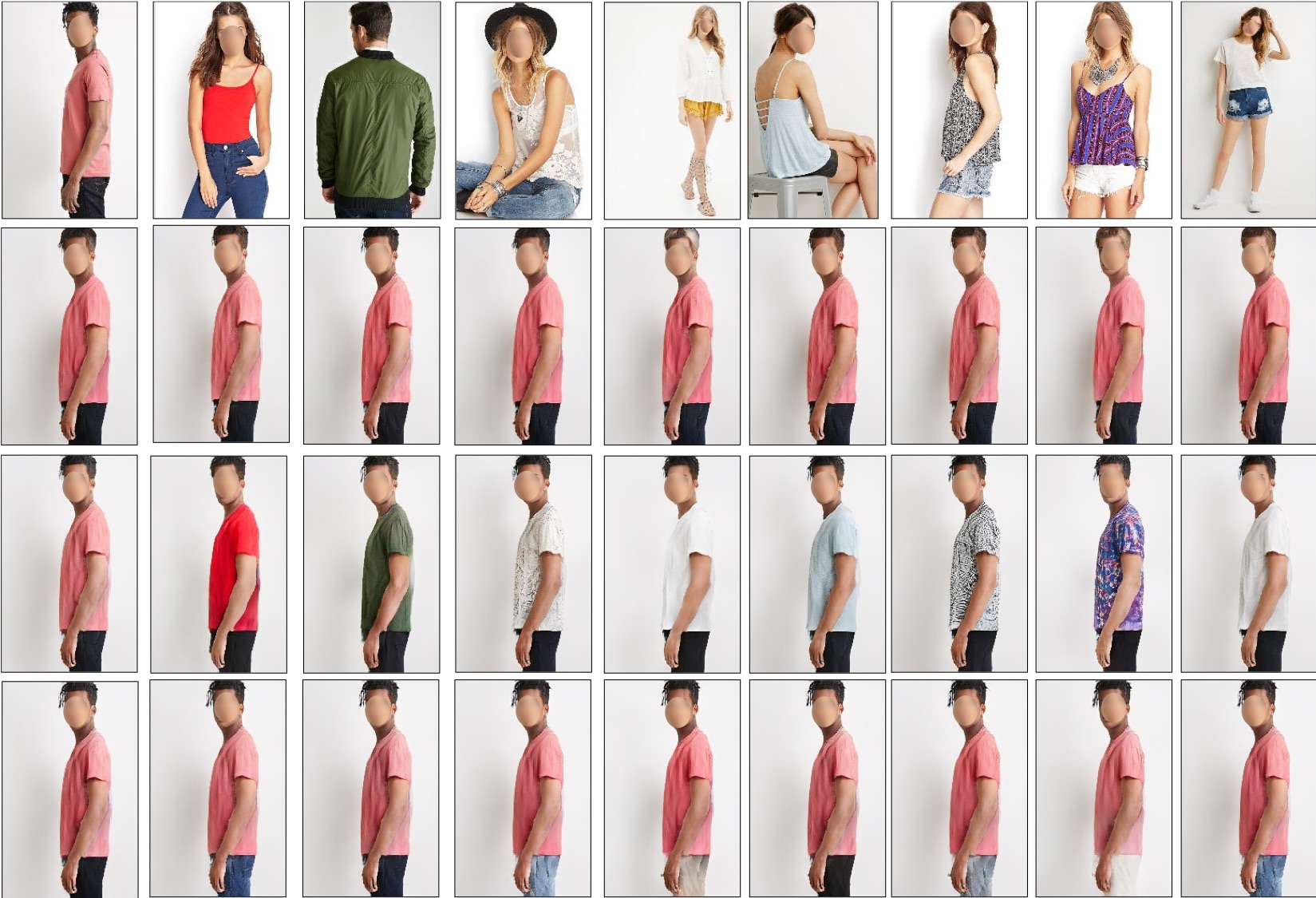} 
  \caption{Replacing the hair, shirt, and pants (DeepFashion). For each target $y$ (row 1), the hair (row 2), shirt (row 3), and pants (row 4), are replaced for the semantic map $s$ of the upper-left person. The EGN and FRN are not used.}
  \label{fig:replace_hair_shirt_pants_df1}
\end{figure}

\clearpage

\begin{figure}[H]
  \centering
 \includegraphics[width=0.9\textwidth]{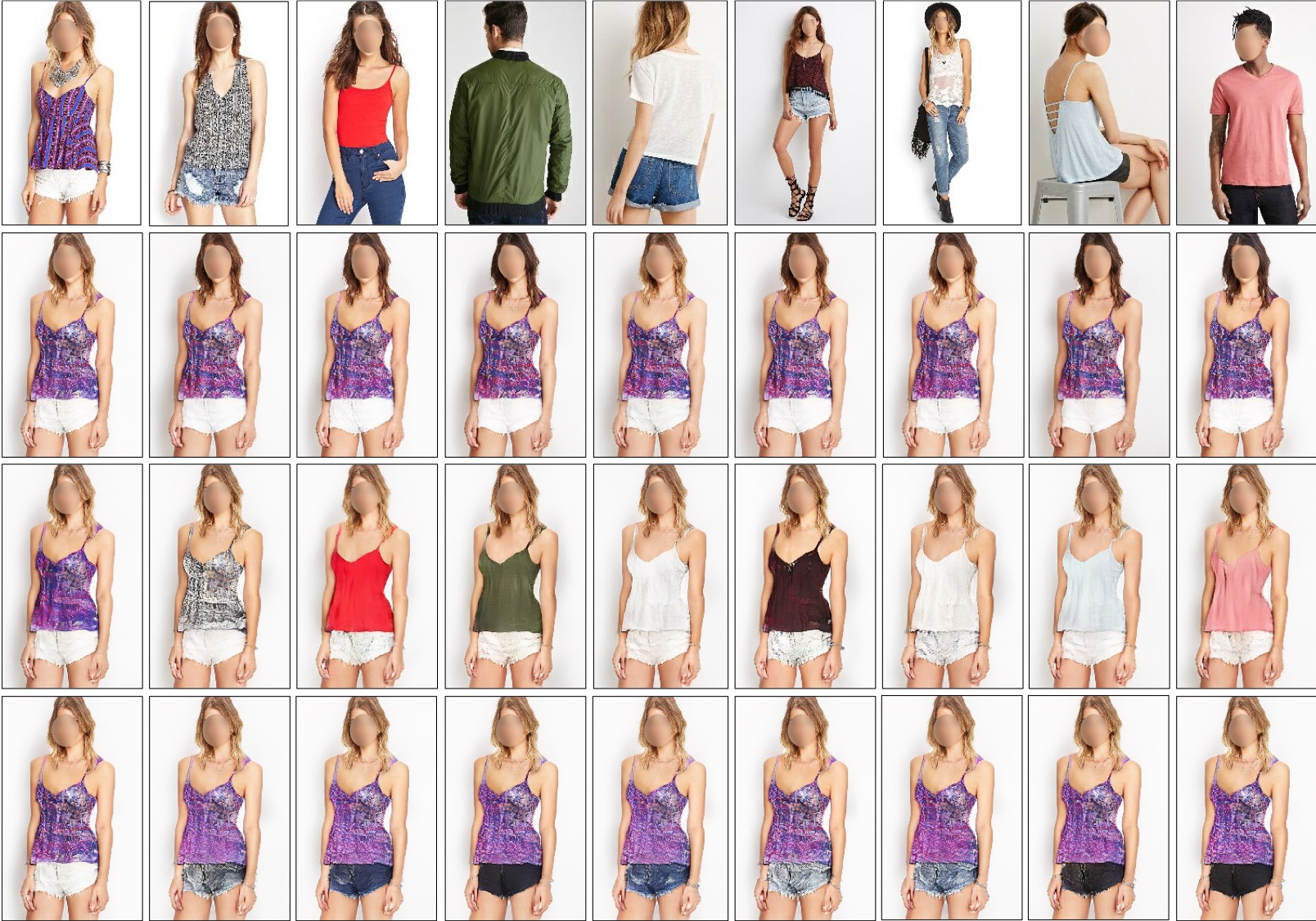} 
  \caption{Replacing the hair, shirt, and pants (DeepFashion). For each target $y$ (row 1), the hair (row 2), shirt (row 3), and pants (row 4), are replaced for the semantic map $s$ of the upper-left person. The EGN and FRN are not used.}
  \label{fig:replace_hair_shirt_pants_df2}
\end{figure}

\clearpage

\begin{figure}[H]
  \centering
 \includegraphics[width=0.9\textwidth]{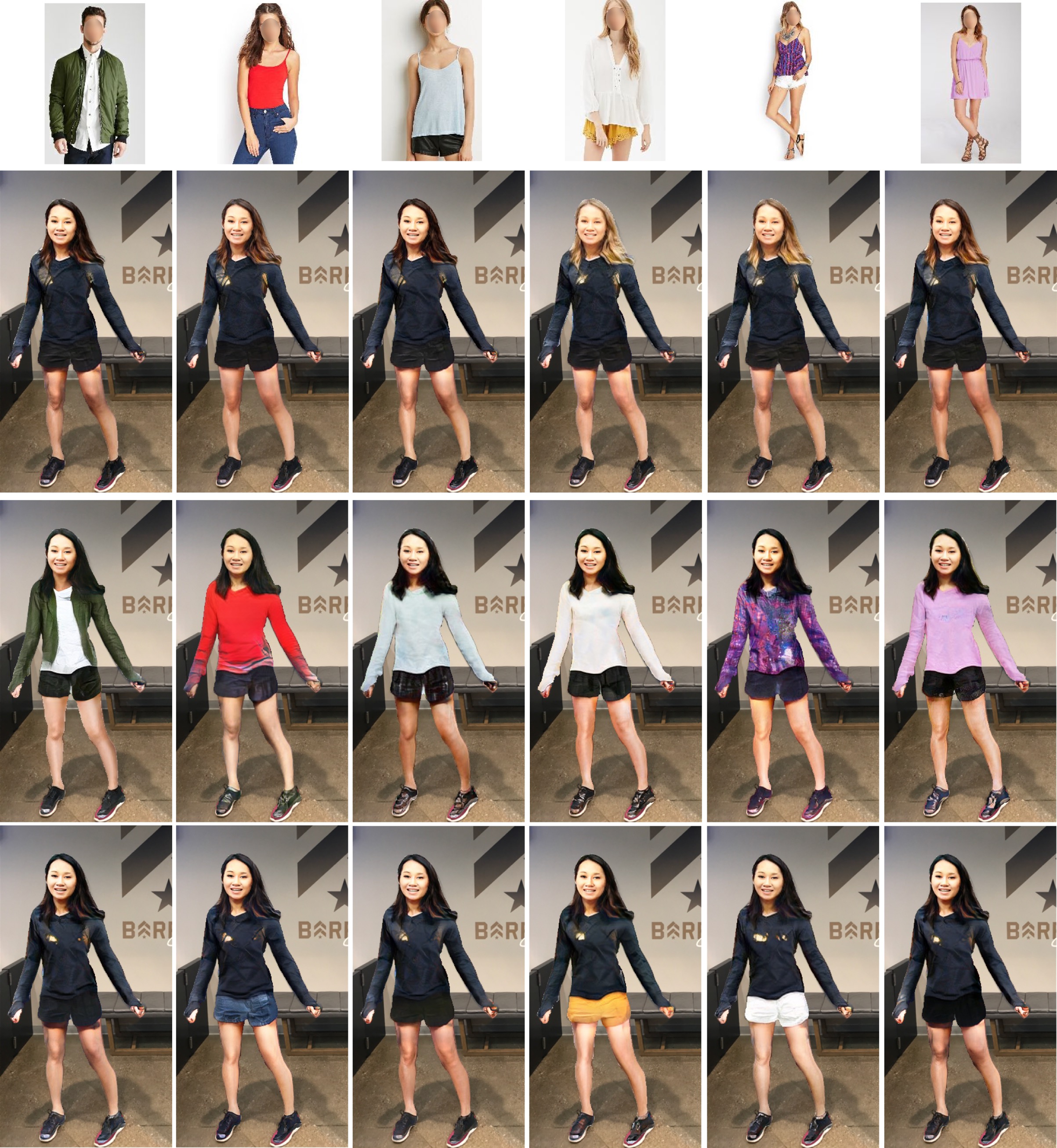} 
  \caption{Replacing the hair, shirt, and pants for high-resolution unconstrained images. For each target $y$ (row 1), the hair (row 2), shirt (row 3), and pants (row 4), are replaced using a chosen semantic map $s$.}
  \label{fig:replace_hair_shirt_pants_mhp}
\end{figure}
\clearpage

\section{Pose-transfer qualitative comparison}

\begin{figure}[H]
  \centering
 \includegraphics[width=0.9\textwidth]{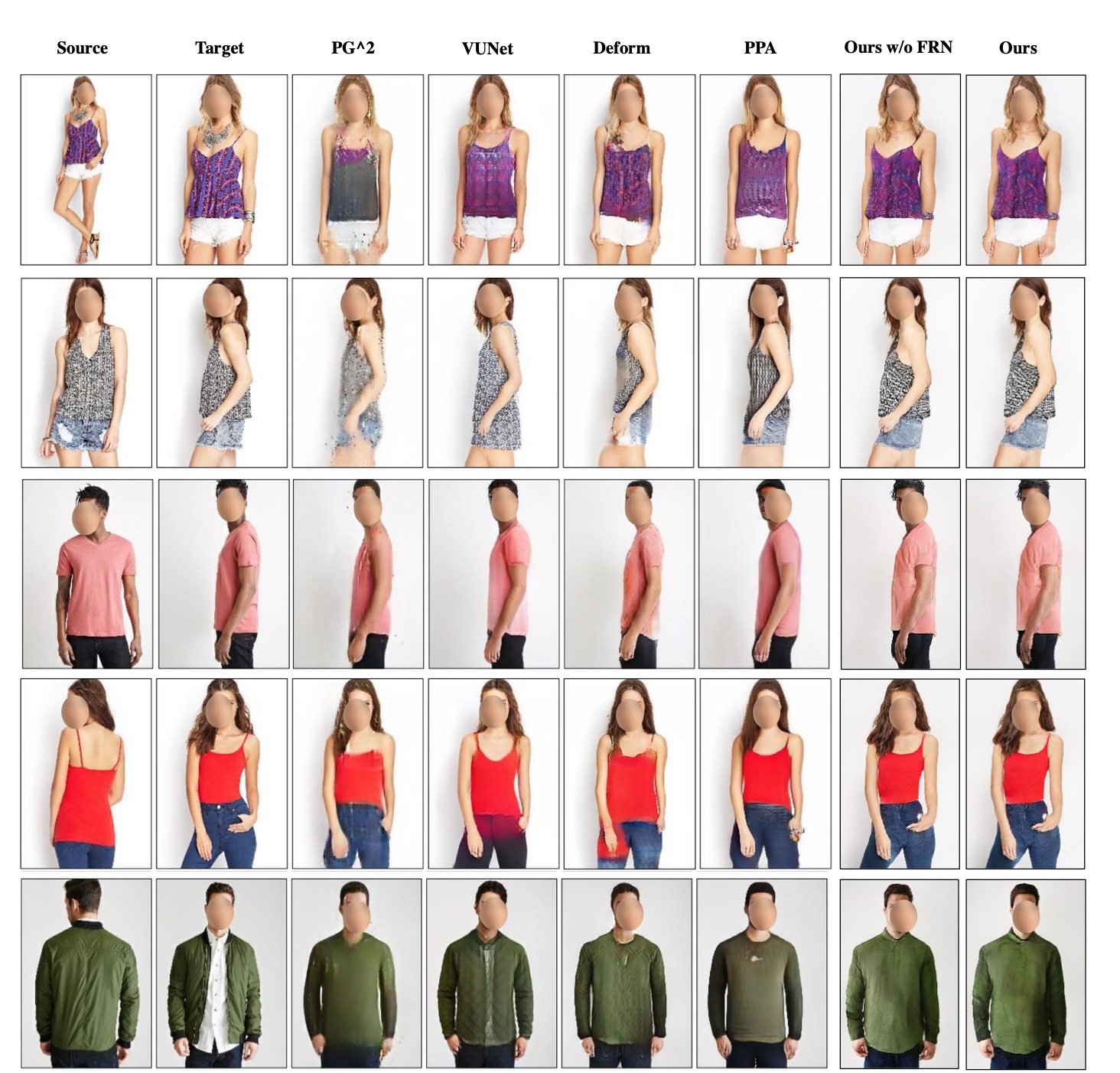} 
  \caption{Comparison of our method on the pose-transfer task. Even without the Face Refinement Network, our method provides photorealistic rendered targets.}
  \label{fig:pose_transfer_compare}
\end{figure}

\clearpage

\section{EGN training samples}
\begin{figure}[H]
  \centering
  \begin{tabular}{c@{~}c@{~}c} 
  \includegraphics[height=7cm]{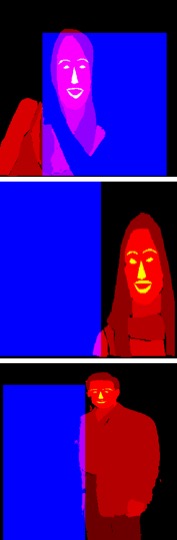} & \includegraphics[height=7cm]{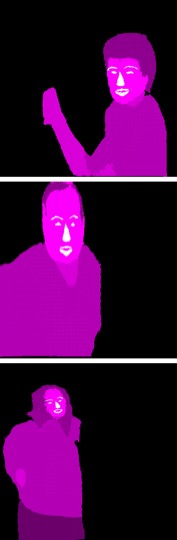} & \includegraphics[height=7cm]{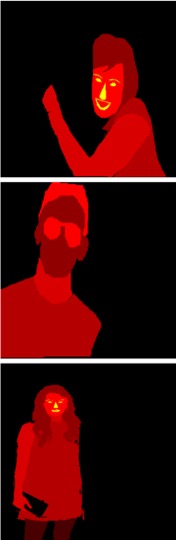} \\
  (a) & (b) & (c)
  \end{tabular}
\caption{Training the Essence Generation Network. Shown for each row are the (a) input semantic map and bounding box, (b) generated semantic map, (c) ground truth semantic map. The scene essence is captured, while the generated semantic map is not identical to the ground truth.}
\label{fig:p2p_train}
\end{figure}


\clearpage
\section{DPBS and DPIS (Python) code}
\begin{lstlisting}[language=Python]
import numpy as np
import os
import cv2

gt_path = 'path/to/ground_truth_densepose'
gen_path = 'path/to/generated_densepose'

read_gen_by_order = True  # Read generated images by order, else by name
n_DP_MAX_IDX = 24  # DensePose generates I with values 0-24

def get_I_iou(img_gt, img_gen, I_idx):
    I_gt = np.zeros_like(img_gt)
    I_gen = np.zeros_like(img_gen)

    I_gt[img_gt == I_idx] = 1  # binarization of the GT image
    I_gen[img_gen == I_idx] = 1  # binarization of the generated image

    I_iou = get_iou(I_gt, I_gen)
    return I_iou

def get_iou(img_gt, img_gen):
    bin_gt = img_gt.copy()
    bin_gen = img_gen.copy()

    bin_gt[bin_gt > 0] = 1  # binarization of the GT image
    bin_gen[bin_gen > 0] = 1  # binarization of the generated image

    bin_union = bin_gt.copy()
    bin_union[bin_gen == 1] = 1  # union over gt and gen (1 where either is present)

    bin_overlap = bin_gt + bin_gen  # overlap of both
    bin_overlap[bin_overlap != 2] = 0  # overlap will be == 2
    bin_overlap[bin_overlap != 0] = 1  # binarization

    union_sum = np.sum(bin_union)
    if union_sum == 0:  # if neither the generated or GT image are present, mask out
        iou = -1
    else:
        iou = np.sum(bin_overlap) / union_sum

    return iou

def get_stats(metric, masked=False):
    if masked:
        return np.ma.mean(metric), np.ma.std(metric), np.ma.median(metric)
    else:
        return np.mean(metric), np.std(metric), np.median(metric)


gt_list = os.listdir(gt_path)  # get ground-truth file names
gt_list.sort()

gen_list = os.listdir(gen_path)  #  get ground-truth files
gen_list.sort()

n_list = len(gt_list)
n_gen = len(gen_list)
if n_list != n_gen:
    print('Error. Ground-truth and generated folders do not contain the same number of images')
    exit(1)
else:
    print('Computing distance metrics over {} images.'.format(n_list))

DPBSs = np.zeros((n_list))  # DensePose Binary Similarity
DPISs = np.zeros((n_list))  # DensePose Index Similarity

for img_idx, filename in enumerate(gt_list):
    img_gt = cv2.imread(os.path.join(gt_path, filename), cv2.IMREAD_UNCHANGED)[:, :, 0]  # DP GT image
    if read_gen_by_order:
        img_gen = cv2.imread(os.path.join(gen_path, gen_list[img_idx]), cv2.IMREAD_UNCHANGED)[:, :, 0]  # DP Generated image read by order
    else:
        img_gen = cv2.imread(os.path.join(gen_path, filename), cv2.IMREAD_UNCHANGED)[:, :, 0]  # DP Generated image read by name

    max_idx = max(np.amax(img_gt), np.amax(img_gen))  #  the max index is taken as the max between the generated and GT
    if max_idx > n_DP_MAX_IDX:
        print('Error. The maximum index value was {}. Should not be over 24'.format(max_idx))
        exit(1)

    DPBSs[img_idx] = get_iou(img_gt, img_gen)  # get DensePose Binary Similarity
    I_ious = np.zeros((n_DP_MAX_IDX))  # DPIS indices per image
    I_mask = np.ones_like(I_ious, dtype=bool)  # masking for DPIS indices per image
    for I_idx in range(1, max_idx + 1):  # iterated over the indices present
        I_ious[I_idx - 1] = get_I_iou(img_gt, img_gen, I_idx)  # index IoU (per body part)
    I_mask[I_ious != -1] = 0  # do not mask IoUs found
    masked_arr = np.ma.array(I_ious, mask=I_mask)  # masked IoUs

    DPISs[img_idx] = np.ma.mean(masked_arr)  # DensePose Index Similarity is calculated over the present indices

    if img_idx % 1000 == 0:
        print('Done with {}/{} images.'.format(img_idx, n_list))
DPBSs_mask = np.zeros_like(DPBSs, dtype=bool)  # masking for DPBS
DPBSs_mask[DPBSs == -1] = 1
masked_DPBSs_arr = np.ma.array(DPBSs, mask=DPBSs_mask)  # masked IoUs

DPBS_mean, DPBS_SD, DPBS_median = get_stats(masked_DPBSs_arr, masked=True)
DPIS_mean, DPIS_SD, DPIS_median = get_stats(DPISs)

print('----------------------------')
print('DPBS')
print('Mean: {}, SD: {}, Median: {}'.format(DPBS_mean, DPBS_SD, DPBS_median))
print('----------------------------')
print('DPIS')
print('Mean: {}, SD: {}, Median: {}'.format(DPIS_mean, DPIS_SD, DPIS_median))
print('----------------------------')

\end{lstlisting}

\end{document}